\documentclass[runningheads]{llncs}
\usepackage{graphicx}
\usepackage{amsmath,amssymb} 
\usepackage{color}
\usepackage{multirow}
\usepackage{makecell}
\usepackage{subcaption}
\usepackage{sidecap}
\sidecaptionvpos{figure}{t}
\usepackage[table]{xcolor}
\usepackage[group-separator={,}]{siunitx}
\usepackage{colortbl,hhline}
\usepackage{enumitem}
\usepackage{stmaryrd}
\usepackage{mathrsfs}
\usepackage[ruled,vlined]{algorithm2e}
\usepackage{bm}
\usepackage{pifont}
\usepackage{amsmath}

\def\etal{\textit{et.al.}}

\begin{document}
\title{Learning without Forgetting for \\3D Point Cloud Objects}
%
\author{Townim Chowdhury\inst{1} \and
Mahira Jalisha\inst{1} \and
Ali Cheraghian\inst{2,3} \and\\
Shafin Rahman\inst{1\thanks{Corresponding author}}\orcidID{0000-0001-7169-0318}}

\institute{North South University, Dhaka, Bangladesh\\ 
\and
Australian National  University,  Canberra, Australia\\
\and
Data61-CSIRO, Australia\\
\email{\{townim.faisal, mahira.jalisha, shafin.rahman\}@northsouth.edu, ali.cheraghian@anu.edu.au}}
\maketitle              

\begin{abstract}
When we fine-tune a well-trained deep learning model for a new set of classes, the network learns new concepts but gradually forgets the knowledge of old training. In some real-life applications, we may be interested in learning new classes without forgetting the capability of previous experience. Such learning without forgetting problem is often investigated using 2D image recognition tasks. In this paper, considering the growth of depth camera technology, we address the same problem for the 3D point cloud object data. This problem becomes more challenging in the 3D domain than 2D because of the unavailability of large datasets and powerful pretrained backbone models. We investigate knowledge distillation techniques on 3D data to reduce catastrophic forgetting of the previous training. Moreover, we improve the distillation process by using semantic word vectors of object classes. We observe that exploring the interrelation of old and new knowledge during training helps to learn new concepts without forgetting old ones. Experimenting on three 3D point cloud recognition backbones (PointNet, DGCNN, and PointConv) and synthetic (ModelNet40, ModelNet10) and real scanned (ScanObjectNN) datasets, we establish new baseline results on learning without forgetting for 3D data. This research will instigate many future works in this area.

\end{abstract}

\section{Introduction}

The advent of deep learning models achieves impressive performance in the image recognition task \cite{simonyan2014very,pointnet2017,pointconv2019}. In a real-life application, a trained system that can classify a given object instance within a fixed number of classes may need to readjust itself to classify a new set of classes in addition to old classes without retraining from scratch. For example, a self-driving car already recognizes street objects (vehicles, traffic lights, etc.). Now, the car manufacturer wants to increase the car's capability in recognizing roadside objects (buildings, trees, etc.) by retraining only on instances of new classes of interest. The main issue of the retraining is the catastrophic forgetting of old class knowledge.  Since this setup does not allow old class instances, the model learns new classes but forgets old ones. Researchers proposed Learning without Forgetting (LwF) methods \cite{lwf2017,icarl2017,hou2019learning,Zhang2020,Rahman_2019_ICCV} to address this problem. Traditionally, this problem has been investigated using 2D image data. This paper explores LwF on 3D point cloud object data.

\begin{figure*}[!t]
\centering
\includegraphics[width=1\textwidth,trim={0cm 0cm 0cm 0cm},clip]{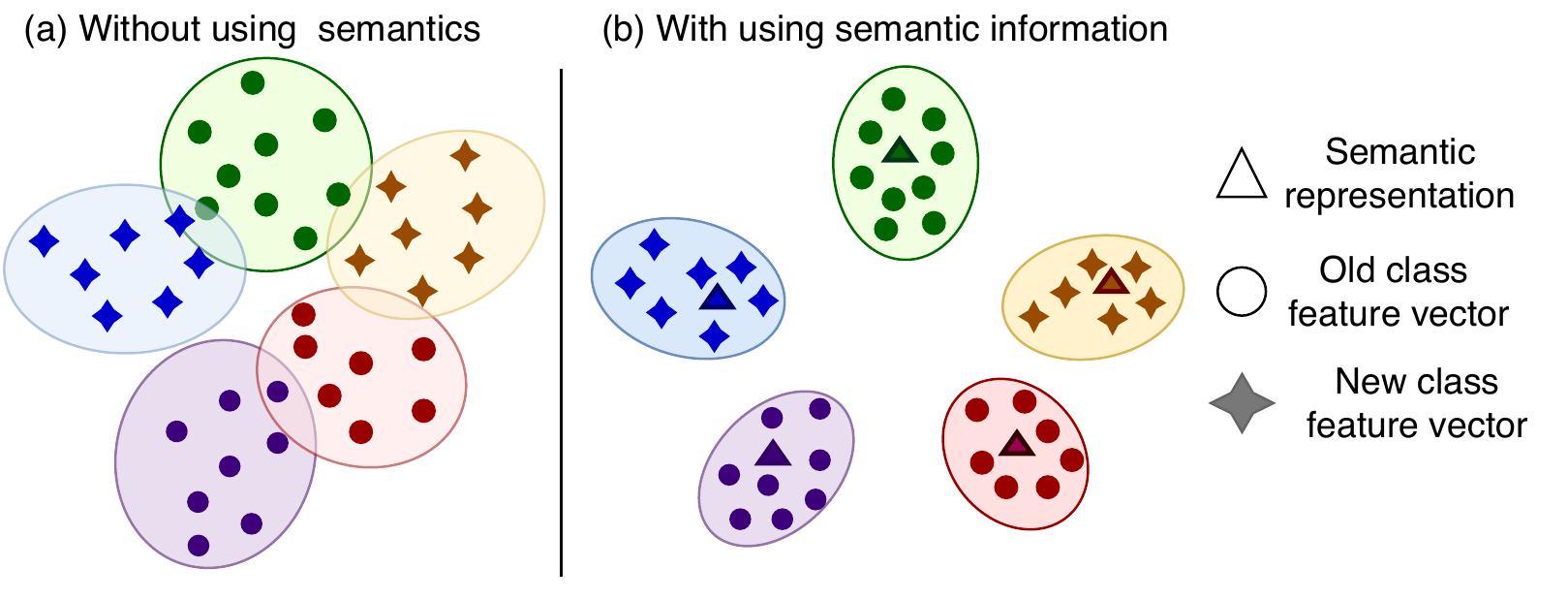}
\caption{Effect of semantic representation while learning without forgetting. (a) Without class semantics, the network tries to form clusters of old and new class instances in feature space. Sometimes clusters overlap with each other because of the lack of class semantics. (b) After using class semantics, old and new class features cluster them around their corresponding class semantics. It helps the cluster separate enough for each other which helps to achieve better performance.}
\label{fig:introduction2}
\end{figure*}

Modern 3D camera technology allows us to capture 3D point cloud data more accessible than ever \cite{9127813}. Now, it is time to adapt 3D point cloud recognition models with LwF capabilities. We identify some key difficulties to address this problem. Firstly, in comparison to image datasets like ImageNet, very large-scale 3D point cloud datasets are not available. 3D datasets usually contain a handful number of classes and instances \cite{modelnet2015,scanobjectnn2019}. Secondly, a typical pre-trained model for a 3D recognition system is not as robust as 2D models because of not being trained on a large dataset \cite{cheraghian2021zero}. Thirdly, 3D point cloud data (especially real scanned objects) contains more noise than 2D image data \cite{scanobjectnn2019}. This paper investigates how far a 3D point cloud recognition model can obtain LwF capabilities considering all difficulties mentioned above.

We first train a 3D point cloud model with instances belonging to a set of pre-defined old classes. Then, we update the trained model using a popular knowledge distillation technique \cite{hinton2015} to address the forgetting problem. Because of the difficulties of 3D data, this approach exhibits a large amount of forgetting of old classes. To minimize forgetting, we employ semantic word vectors of classes inside the network pipeline \cite{10.1007/978-3-030-69535-4_6,Ali_2021_CVPR,Semantic_Relation_2021_CVPR}. During both new and old task training, the network tries to align point cloud features to their corresponding semantics. The class semantics encodes similarities and dissimilarities of different objects from the natural language domain. The network learns to project new instance features around the previously obtained and fixed semantic vectors while learning new classes. By performing feature-semantic alignment in both old and new tasks, the network forgets less than the traditional semantic embedding less method. For example, during the old model training, the model learns to classify `bed' via its semantic (like isFurniture, isIndoor) representation. Later, during the new model training, the model could not see `bed', but it observes similar classes (like sofa, chair, table with shared `bed' semantics) that helps not to forget about `bed' knowledge. Experimenting on ModelNet40 \cite{modelnet2015}, ScanObjectNN \cite{scanobjectnn2019}, MIT Scenes \cite{scene2009}, and CUB \cite{cub2011} datasets, we show that our proposed method outperforms traditional knowledge distillation methods in both 3D and 2D data cases. The contributions of this paper are summarized below:
\begin{itemize}
    \item To the best of our knowledge, we are the first to experiment learning without forgetting on 3D point object cloud data. 
    
    \item Our method applies knowledge distillation to restore previously gained experience of the old mode and minimize catastrophic forgetting while learning a set of new classes. In addition, we investigate the advantage of semantic word vectors in the network distillation process.
    
    \item We experiment on both 3D synthetic (ModelNet10, ModelNet40 \cite{modelnet2015}) and real scanned (ScanObjectNN \cite{scanobjectnn2019}) point cloud objects and 2D image datasets (MIT Scenes \cite{scene2009}, CUB \cite{cub2011}), establishing the robustness of the method.
\end{itemize}

\section{Related Works}
\textbf{3D Point Cloud Architecture:} There are two streams of works for 3D point cloud classification: feature-based and end-to-end approaches. Feature-based methods mostly use Multi-view representation and Volumetric CNNs. Multi-view representation methods \cite{su2015multi,qi2016volumetric,yang2019learning} convert 3D point cloud into 2D images, which are then classified using 2D convolutional networks. Volumetric CNNs \cite{maturana2015voxnet,wang2017cnn} project point cloud objects on a volumetric grid or a set of octrees. Then, they apply a computationally expensive 3D convolutional neural network. The main drawback of feature-based methods is that they do not work directly on the raw point cloud. End-to-end approaches like PointNet \cite{pointnet2017}, PointNet++\cite{qi2017pointnet++} use raw point cloud data as input to multi-layer perceptron networks followed by maxpooling layers. Several other works \cite{pointconv2019,komarichev2019cnn,li2018pointcnn} apply improved convolution operation on point cloud objects. Moreover, \cite{simonovsky2017dynamic,wang2019dynamic} use Graph neural networks to extract features from 3D point clouds. In this paper, we build our model on several end-to-end architectures.

\noindent \textbf{Learning Without Forgetting:} Many methods have been proposed to solve the catastrophic forgetting problem \cite{catastrophic1989,empirical2014,aljundi2017expert}. Exemplar-free methods \cite{lwf2017,aljundi2018memory,Zhang2020} do not require any samples from base/old task. Li \etal{} \cite{lwf2017} proposed to use Hinton's \cite{hinton2015} knowledge distillation loss to preserve old task's knowledge in 2D images, but the domain shift between tasks makes this method weak. Rehearsal methods \cite{icarl2017,castro2018,hou2019learning} keep a small number of exemplars from the old task. Rebuffi \etal{} \cite{icarl2017} first introduced replay-based method with bounded memory, but it fails to represent the main distribution of old task when there is a lot of variations. The Pseudo-rehearsal process used in \cite{dgr,wu2018memory,ostapenko2019learning} learns to produce examples from the old task. Some methods \cite{aljundi2018memory,mallya2018packnet} minimizes additional parameters to solve the problem of model expansion. All of the approaches mentioned above have proposed solutions to the catastrophic forgetting of 2D image data. Our method is the first to use knowledge distillation to address LwF of 3D data.

\noindent \textbf{Word embedding for catastrophic forgetting:} The use of semantic representation to prevent catastrophic forgetting is relatively new \cite{10.1007/978-3-030-69535-4_6,Ali_2021_CVPR,Semantic_Relation_2021_CVPR}. Such approaches explore the semantic relation between old and new classes to reduce the forgetting of old classes while training new classes. Zhu \etal{} \cite{Semantic_Relation_2021_CVPR} suggested using semantic representation to train the object detection model by projecting the feature vector into the semantic space. Similarly, Rahman \etal{} \cite{10.1007/978-3-030-69535-4_6} proposed to use semantic representations of class labels as anchors in the prediction space for not forgetting the acquired knowledge of old classes. Cheraghian \etal{} \cite{Ali_2021_CVPR} proposed a knowledge distillation strategy by using semantic representation as an auxiliary knowledge. Even though semantic representation has yielded promising results, the experiments are limited to 2D data. In this paper, we use word vectors in knowledge distillation for 3D point clouds object classification.

\section{Method}


\noindent\textbf{Problem Formulation:} Assume, we have a set of old, $\mathcal{Y}^{o}$, and a set of new, $\mathcal{Y}^{n}$, classes, where, $\mathcal{Y}^{o} \cap \mathcal{Y}^{n}=\varnothing$,  $|\mathcal{Y}^{o}|=O$ and $|\mathcal{Y}^{n}|=N$. The 3D point cloud recognition model initially observes $\mathcal{Y}^{o}$ classes and gets trained to classify only old classes. Later, $\mathcal{Y}^{n}$ classes are added to the model to update previous training. Suppose,  a 3D point cloud input sets, $\mathcal{X} = \{x^i\}_{i=1}^n$ for $x^i\in\mathbb{R}^3$, can get a label from either old $\mathcal{Y}^{o}$ or new $\mathcal{Y}^{n}$ classes. Additionally, there is a set of $d$-dimensional semantic class embeddings for each of the old and new classes, denoted as $\mathcal{E}^{o} \in \mathbb{R}^{d\times O}$ and $\mathcal{E}^{n} \in \mathbb{R}^{d\times N}$, respectively. We define the old set as $\mathcal{D}^{o}=\left \{ \mathcal{X}^{o}_{i},y^{o}_{i},e^{o}_{i} \right \}_{i=1}^{N_{o}}$, where, $\mathcal{X}^{o}_{i}$ is the $i$-th point cloud object belonging to old set with the class label $y^{o}_{i}$, and the class embedding $e^{o}_{i}$, and $N_{o}$ is the number of old class instances. Similarly, there is a set for new classes $\mathcal{D}^{n}=\left \{ \mathcal{X}^{n}_{i},y^{n}_{i},e^{n}_{i} \right \}_{i=1}^{N_{n}}$, where, $\mathcal{X}^{n}_{i}$ is the $i$-th point cloud object having the class label $y^{n}_{i}$, and the class embedding $e^{n}_{i}$, and $N_{n}$ is the number of new class instances. We build a 3D point cloud object recognition model (termed as old model) using $\mathcal{D}^{o}$ set. Then, we aim to update the same model (termed as new model) using only newly available $\mathcal{D}^{n}$ data that can predict a class label for a test sample belonging to either old or new sets, \textit{i.e.}, $\mathcal{X} \in \mathcal{D}^{o} \cup \mathcal{D}^{n}$. We assume the model has prior knowledge about the test sample during inference, whether it belongs to old or new classes.

\noindent\textbf{Main challenges:} While updating the model with new data, $\mathcal{D}^{n}$, the model gradually forgets the old training (done on $\mathcal{D}^{o}$) because of the restriction of not using old class instances. Previous works address this problem with 2D image data. In this paper, we address the same problem on 3D point cloud objects. Due to the unavailability of large-scale datasets and pre-trained models, the problem becomes more complex in the 3D than 2D domain.

\begin{figure*}[!t]
\centering
\includegraphics[width=1\textwidth,trim={1.2cm 0cm 0cm 0cm},clip]{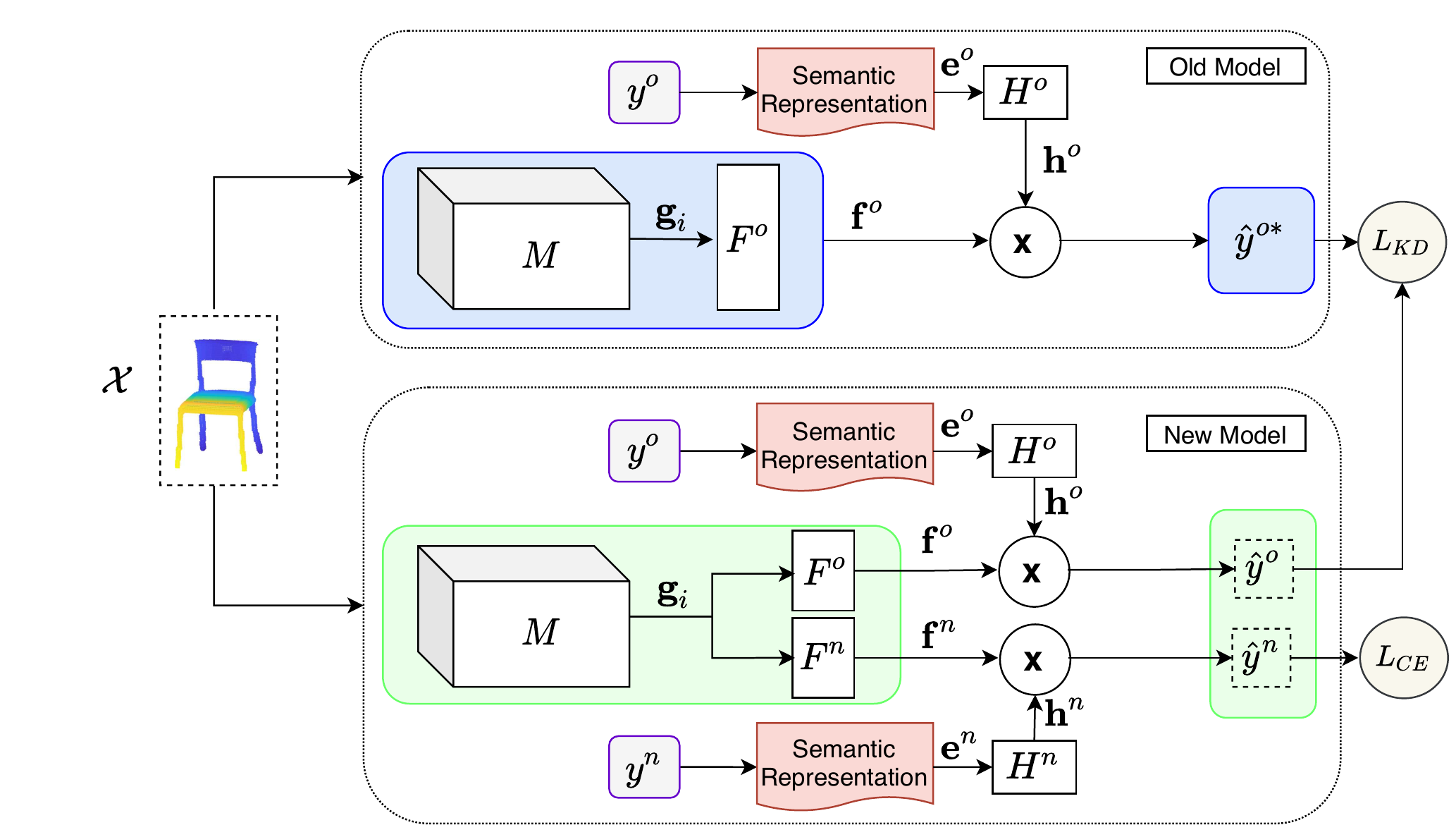}
\caption{Our proposed architecture.  We train the old model using the cross-entropy loss $L_{CE}$. Then, we build a new model by modifying a copy of the trained old model. Both cross-entropy, $L_{CE}$ and knowledge distillation, $L_{KD}$ losses are used to train the new model. The new model can classify both old and new classes.
}
\label{fig:our_approach}
\end{figure*}

\subsection{Model Overview}
\label{sec:overview}

Our proposed method is shown in Fig~\ref{fig:our_approach}, which includes old and new models. The new model is the updated variant of the old model to accommodate new classes. Both old and new models are presented together because, during the training of the new model, we use the output of the old model.
For both models, the point cloud input $\mathcal{X}$ is fed into the backbone $M$, which can be any point cloud architecture (PointNet, DGCNN, PointConv etc.), to extract feature input,\textit{i.e.}, $\textbf{g} \in \mathbb{R}^{m}$. Additionally, a semantic representation unit is employed to generate class embedding,\textit{i.e.}, $\textbf{e} \in \mathbb{R}^{d}$, given class label. 
While training the old model using old classes, the feature input $\textbf{g}$ and the class embedding $\textbf{e}^o$ are mapped into a common $k$-dimensional space using projection functions $F^{o}$ and $H^{o}$, respectively. The new feature representations for the point cloud feature and the class embedding are $\textbf{f}^{o} \in \mathbb{R}^{k}$ and $\textbf{h}^{o} \in \mathbb{R}^{k}$, respectively. Finally, dot multiplication between $\textbf{f}^{o}$ and $\textbf{h}^{o}$ form the output $\hat{\textbf{y}}^{o}$ for the old classes. A cross-entropy loss, $L_{CE}$, is adopted to train the model for the old classes. 
While updating the same model with the new classes, we add a parallel pipeline from the output of backbone $M$. Two projection modules $F^{n}$ and $H^{n}$ are added to map features of new classes $M(\mathcal{X}^{n})$ and class embedding $\textbf{e}^{n}$ into the common $k$-dimensional space. The new representations of feature and class semantics are $\textbf{f}^{n}$ and $\textbf{h}^{n}$, respectively. At the end, $\textbf{f}^{n}$ is dot-multiplied with $\textbf{h}^{n}$ to generate output, $\hat{\textbf{y}}^{n}$ for the new classes. In order to prevent forgetting of the old classes, a knowledge distillation loss, function, $L_{KD}$, \cite{hinton2015} is employed between output of the old and new models.

\subsection{Training and Inference} \label{sec:training}

We train the proposed architecture with two stages: old and new model training. Unlike traditional approaches for learning without forgetting (LwF) \cite{lwf2017}, both stages use semantic word vectors of classes to remember past knowledge. 

\noindent\textbf{Training old model:} At the first stage of training, we learn the old model using the training data of $\mathcal{D}^{o}=\left \{ \mathcal{X}^{o}_{i},y^{o}_{i},e^{o}_{i} \right \}_{i=1}^{N_{o}}$ employing a cross-entropy loss. Unlike 2D image cases, we perform this training from scratch because no pre-trained model is available to initialize the weights of the backbone, $M$. The output of the old model for the $i$th 3D point cloud instance is
\begin{equation}
\mathbf{\hat{y}}^{o*}_i = F^o(\textbf{g}_i;\theta_{1})\: . \:H^o(\textbf{e}_{i};\theta_{2})^T
\label{eq:oldmodel}
\end{equation}
where, $\theta_{1}$ and $\theta_{2}$ are learnable weights associate with $F^o$ and $H^o$ layers, respectively. After finishing the training, the old model can classify the old set of classes, $\mathcal{Y}^{o}$. This old model remains frozen during the second stage training.

\noindent\textbf{Training new model:} We build a new model during the second training stage by updating a copy of the old model, which is trained in the previous stage. This new model gives predictions for both old and new classes. But, we are not allowed to cannot any old class instances during training the new model. We add $F^{n}$ and $H^{n}$ layers to $F^{o}$ and $H^{o}$ layers. Only the training data of new classes $\mathcal{D}^{n}=\left \{ \mathcal{X}^{n}_{i},y^{n}_{i},e^{n}_{i} \right \}_{i=1}^{N_{n}}$ is used to train the new model. Similar to Eq. \ref{eq:oldmodel}, both pipeline of the new model can produce output for old $\mathbf{\hat{y}}^{o}_i$ and new $\mathbf{\hat{y}}^{n}_i$ classes. 
\begin{align}
\mathbf{\hat{y}}^{o}_i = F^o(\textbf{g}_i;\theta_{1})\:.\:H^o(\textbf{e};\theta_{2})^T, \quad \quad \quad
\mathbf{\hat{y}}^{n}_i = F^n(\textbf{g}_i;\theta_{3})\:.\:H^n(\textbf{e};\theta_{4})^T
\label{eq:newmodel}
\end{align}
where, $\theta_{3}$ and $\theta_{4}$ are weights associated with $F^n$ and $H^n$ layers, respectively. Among all trainable weights of new model,  $\theta_{1}$ and $\theta_{2}$ are initialized from the old model but $\theta_{3}$ and $\theta_{4}$ are trained from the scratch. While forwarding an input 3D point cloud object $x_i\in\mathcal{X}$, old model outputs $\mathbf{\hat{y}}^{o*}_i$ for old classes and new model outputs $\mathbf{\hat{y}}^{o}_i$ and $\mathbf{\hat{y}}^{n}_i$ for old and new classes, respectively.

We calculate a traditional cross-entropy loss $L_{CE}$ between $\mathbf{\hat{y}}^{n}_i$ and ground-truth $y^{n}_{i}$. This loss is used to learn new classes. Additionally, using old class outputs from old $\mathbf{\hat{y}}^{o*}_i$ and new $\mathbf{\hat{y}}^{o}_i$ model, we calculate a knowledge distillation \cite{hinton2015} loss $L_{KD}$. This loss is employed to prevent the forgetting of the old classes. Unlike the traditional $L_{KD}$, we consider class semantics in the pipeline, which further helps the LwF process. The entire loss ($L$) to train this model is
\begin{equation}\label{eq:overall_loss}
\begin{split}
    L = L_{CE}+\lambda L_{KD}
\end{split}
\end{equation}
where, hyperparameter $\lambda$ controls the contribution of $L_{KD}$. To calculate $L_{CE}$, we use negative log likelihood loss common in 3D backbones. To calculate $L_{KD}$, we record the output $\hat{\textbf{y}}^{o\ast}$ from old model for new class dataset's 3D point clouds $\mathcal{X}^n$. The equations for $L_{CE}$ and $L_{KD}$ are:
\begin{align}
L_{CE} = -\frac{1}{N} \sum_{i=1}^{N}\mathbf{y}^{n}_{i}log(\mathbf{\hat{y}}^{n}_i), \quad \quad \quad
    L_{KD} = -\frac{1}{N} \sum_{i=1}^{N} \omega(\hat{\textbf{y}}^{o\ast }_{i};\tau)log(\omega(\hat{\textbf{y}}^{o}_{i};\tau))
\label{eq:kd_loss}
\end{align}
where, 
\(\tau\) is the temperature  and $\omega(y_i;\tau)  = \frac{exp(y_i/\tau)}{\sum_{j}^{}exp(y_{j}/\tau)}$ is the softmax function.

\noindent\textbf{Inference:} For any test instance, a forward pass to the new model calculates old $\mathbf{\hat{y}}^{o}_i$ and new $\mathbf{\hat{y}}^{n}_i$ class scores. We classify old and new classes by selecting the maximum score from $\mathbf{\hat{y}}^{o}_i$ and $\mathbf{\hat{y}}^{n}_i$, respectively.

\section{Experiment}

\noindent\textbf{Dataset:} We evaluate our method on 3D datasets, ModelNet10, ModelNet40 \cite{modelnet2015}, ScanObjectNN \cite{scanobjectnn2019} and two 2D datasets, MIT Scenes \cite{scene2009}, CUB \cite{cub2011}. 
For the 3D experiment, we use two different settings related to synthetic and real scanned point cloud data. The synthetic experiment, ModelNet40 $\rightarrow$ ModelNet10 setting use 30 classes of ModelNet40 as old and non-overlapped 10 classes of ModelNet10 as new classes. The real scanned object experiment, ModelNet40 $\rightarrow$ ScanObjectNN use 26 classes of ModelNet40 as old and 11 classes of ScanObjectNN as new classes. Both of these setups were previously introduced in \cite{cheraghian2021zero}.
For the 2D experiment, Scenes $\rightarrow$ CUB considers 67 classes of MIT Scenes as old and 200 classes of CUB as new. In another setup, 150 and 50 classes of CUB dataset are used as old and new, respectively. These setups are proposed in \cite{lwf2017,xian2018zero}. The statistics of train test instances are summarized in Table \ref{table:dataset}.

\begin{table}[!t] \centering \small
\centering
\scalebox{.92}{\begin{tabular*} {\linewidth}{@{\extracolsep{\fill}}cccccc}
\hline   
Dataset & Settings & Task & \# Classes & \# Train & \# Test \\ 
\hline
\multirow{4}{*}{3D} & \multirow{2}{*}{ModelNet40 $\rightarrow$ ModelNet10} & old & 30 & 5852 & 1560\\
& & new & 10 & 3991 & 908\\
\cline{2-6}
 & \multirow{2}{*}{ModelNet40 $\rightarrow$ ScanObjectNN} & old & 26 & 4999 & 1496\\ 
&  & new & 11 & 1496 & 475\\ 
\hline
\multirow{4}{*}{2D} & \multirow{2}{*}{Scenes $\rightarrow$ CUB} & old & 67 & 5360 & 1340\\
& & new & 200 & 5994 & 5794\\
\cline{2-6}
 & \multirow{2}{*}{CUB} & old & 150 & 4495 & 4326\\
&  & new & 50 & 1499 & 1468\\ 
\hline
\end{tabular*}}
\caption{Statistics of training and testing instances used in different experiments.}
\label{table:dataset}
\end{table}

\noindent\textbf{Semantic embedding:} For semantic representation of classes, we use 300 dimensional word2vec (w2v) \cite{Mikolov_NIPS_2013} and GloVe (glo) \cite{pennington2014glove} word vectors. The word vector models are usually trained on unannotated text corpus. Unless explicitly mentioned all performance in this paper are with word2vec vectors.

\noindent\textbf{Evaluation protocol:} We evaluate our method using top-1 accuracy. We calculate the old model's accuracy as $Acc_{o}^*$. Similarly, we calculate $Acc_{o}$ and $Acc_{n}$ to represent performance of old and new classes, respectively using the final model. To measure the extent of forgetting, we calculate, $\Delta = \frac{Acc_{o}^* - Acc_{o}}{Acc_{o}^*} \times 100\%$. A lower $\Delta$ indicates less forgetting of the new model.

\noindent\textbf{Validation strategy:} We further randomly divide the set of old classes into val-old and val-new classes for validation experiment. In the ModelNet40 $\rightarrow$ ModelNet10 and ModelNet40 $\rightarrow$ ScanObjectNN experiments, we choose 24 and 20 classes from old classes, respectively as val-old and the rest of the classes as val-new to find values for hyperparameters. We choose $\lambda=3$ and $\tau=3$ for our 3D experiments by performing a grid search within the range $(0,3]$. 

\noindent\textbf{Implementation details\footnote{Codes are available at: \url{https://github.com/townim-faisal/lwf-3D}}:} We use PointNet \cite{pointnet2017}, PointConv \cite{pointconv2019}, DGCNN \cite{wang2019dynamic} as 3D point cloud backbone and VGG16 \cite{simonyan2014very} (pretrained on Imagenet \cite{imagenet2015}) as 2D image backbone to obtain feature vector. For feature vector projection layers, we use two fully connected layers (512, 256) and (1024, 512) with ReLU activations in 3D and 2D experiments, respectively. For 3D and 2D experiments, we use one fully connected layer of size 256 and 512 with ReLU in the projection layer of semantic representation. In all experiments, we use the Adam optimizer with a learning rate of 0.0001 and batch sizes of 32 during training. We implement our work using the \textit{PyTorch} framework.

\noindent\textbf{Compared methods:} In this paper, we compare the results of the following methods.
\textbf{\textit{(a)}} Baseline-1: A backbone model is trained using the instances of old classes. Then, the trained backbone is further fine-tuned using new class instances only.
\textbf{\textit{(b)}} LwF \cite{lwf2017}: The backbone training is same as Baseline-1. Then, the fine-tuning on new class samples uses a knowledge distillation loss \cite{hinton2015} not to forget the old class knowledge.
\textbf{\textit{(c)}} Baseline-2: This method is an intermediate stage of our proposed approach. We first train the old model of Fig. \ref{fig:our_approach} using semantic word vector information inside the architecture. But, it does not have any fine-tuning stage. This performance can be considered zero-shot learning \cite{cheraghian2021zero,rahman2020IJCV} results because it treats new classes as unseen. This method can classify new (unseen) classes without having trained on new instances.
\textbf{\textit{(d)}} Ours: This is our final recommendation as described in Sec. \ref{sec:overview} and \ref{sec:training}. On top of Baseline-2 training, it contains fine-tuning on new class instances.

\begin{table}[t]
\newcolumntype{C}[1]{>{\centering\arraybackslash}p{#1}}
\begin{minipage}{.5\linewidth}
\caption*{ModelNet40 $\rightarrow$ ModelNet10}
\centering \small
\scalebox{0.95}{\begin{tabular}{C{1.5cm}C{1cm}C{1cm}C{1cm}C{1cm}}
\hline
Method & $Acc_o^{\ast}\uparrow$ & $Acc_o\uparrow$ & $Acc_n\uparrow$ & $\Delta\downarrow$\\
\hline
Baseline-1 & 89.2 & 41.5 & 90.2 & 53.5\\
LwF \cite{lwf2017} & 89.2 & 83.6 & 89.3 & 6.2\\ 
Baseline-2 & 89.4 & - & 22.8 & -\\
Ours & \textbf{89.4} & \textbf{84.4} & \textbf{90.4} & \textbf{5.5}\\ 
\hline
\end{tabular}}
\end{minipage}
\begin{minipage}{.5\linewidth}
\caption*{ModelNet40 $\rightarrow$ ScanObjectNN}
\centering \small
\scalebox{0.95}{\begin{tabular}{C{1.5cm}C{1cm}C{1cm}C{1cm}C{1cm}}
\hline
Method & $Acc_o^{\ast}\uparrow$ & $Acc_o\uparrow$ & $Acc_n\uparrow$ & $\Delta\downarrow$\\
\hline
Baseline-1 & 89.8 & 51.0 & \textbf{76.9} & 43.3\\
LwF \cite{lwf2017} & 89.8 & 81.3 & 73.7 & 9.5\\ 
Baseline-2 & 89.9 & - & 21.5 & -\\ 
Ours & \textbf{89.9} & \textbf{86.2} & 74.6 & \textbf{4.1}\\ 
\hline
\end{tabular}}
\end{minipage} 
\caption{3D data experiment using PointNet. $\uparrow$ ($\downarrow$) means higher (lower) is better.}
\label{table:overall_experiment_pointnet}
\end{table}

\begin{table}[!t]
\centering \small
\centering
\vspace{-1em}\begin{tabular*} {\linewidth}{@{\extracolsep{\fill}}ccccc|cccc}
\hline
\multirow{2}{*}{Backbone} & \multicolumn{4}{c}{ModelNet40 $\rightarrow$ ModelNet10} & \multicolumn{4}{c}{ModelNet40 $\rightarrow$ ScanObjectNN}\\
\cline{2-9}
& $Acc_o^{\ast}\uparrow$ & $Acc_o\uparrow$ & $Acc_n\uparrow$ & $\Delta\downarrow$ & $Acc_o^{\ast}\uparrow$ & $Acc_o\uparrow$ & $Acc_n\uparrow$ & $\Delta\downarrow$\\
\hline
PointNet \cite{pointnet2017} & 89.4 & 84.4 & 90.4 & 5.5 & 89.9 & \textbf{86.2} & 74.6 & \textbf{4.1}\\
PointConv \cite{pointconv2019} & 90.5 & 86.2 & 87.8 & \textbf{4.8} & 90.2 & 73.4 & 66.6 & 18.6\\
DGCNN \cite{wang2019dynamic} & \textbf{91.5} & \textbf{87.1} & \textbf{93.4} & 4.9 & \textbf{91.6} & 71.8 & \textbf{75.0} & 21.6\\
\hline
\end{tabular*}
\caption{Ablation study on varying 3D point cloud backbone.}\vspace{-2em}
\label{table:varying_backbone}
\end{table}

\subsection{3D point cloud experiments}

\noindent\textbf{Overall results:} Table \ref{table:overall_experiment_pointnet} shows the overall results using two settings, ModelNet40 $\rightarrow$ ModelNet10 and ModelNet40 $\rightarrow$ ScanObjectNN. Our observations are as follows.
\textbf{\textit{(1)}} Baseline-1 gets the worst results in forgetting issue showing high $\Delta$ values because the fine-tuning for the new model does not consider about old classes. High and low value in $Acc_n$ and $Acc_o$, respectively tells that this method learns new classes but forgets the old classes significantly.
\textbf{\textit{(2)}} LwF \cite{lwf2017} obtains better results on forgetting issue (lower $\Delta$ values) than Baseline-1 because this method apply a knowledge distillation loss not to forget old classes.
\textbf{\textit{(3)}} Baseline-2 shows the performance after old class training using our method. Without receiving training on new classes, this model can still classify new classes considering those as unseen class. Although no forgetting occurred in this case, there is no balance of old and new class performance.
\textbf{\textit{(4)}} Ours result makes a nice balance of old and new accuracy with maintaining minimal forgetting.
\textbf{\textit{(5)}} Although both settings achieve similar results ($Acc_o$) in old classes across methods, ModelNet40 $\rightarrow$ ScanObjectNN gets less accuracy on new classes ($Acc_n$) than ModelNet40 $\rightarrow$ ModelNet10. The reason is that ScanObjectNN classes (new) are real-scanned 3D objects with higher noise than synthetic data.

\begin{table}[t]
\newcolumntype{C}[1]{>{\centering\arraybackslash}p{#1}}

\begin{minipage}{.5\linewidth}
\caption*{(a) Using PointNet backbone}
\centering \small
\scalebox{0.85}{\begin{tabular}{C{2.4cm}C{.8cm}C{.8cm}C{.8cm}C{.8cm}C{.8cm}}
\hline
Settings & Word & $Acc_o^{\ast}\uparrow$ & $Acc_o\uparrow$ & $Acc_n\uparrow$ & $\Delta\downarrow$\\
\hline
ModelNet40 & glo & 88.2 & 78.8 & \textbf{90.6} & 10.6\\
$\rightarrow$ModelNet10 & w2v & \textbf{89.4} & \textbf{84.4} & 90.4 & \textbf{5.5}\\ 
\hline
ModelNet40 & glo & 89.7 & 85.2 & 70.9 & 5.0\\
$\rightarrow$ScanObjectNN & w2v & \textbf{89.9} & \textbf{86.2} & \textbf{74.6} & \textbf{4.1}\\ 
\hline
\end{tabular}}
\end{minipage}
\begin{minipage}{.5\linewidth}
\caption*{(b) Using VGG16 backbone}
\centering \small
\scalebox{0.85}{\begin{tabular}{C{1.9cm}C{1.4cm}C{.8cm}C{.8cm}C{.8cm}C{.8cm}C{.8cm}}
\hline
Settings & Method & $Acc_o^{\ast}\uparrow$ & $Acc_o\uparrow$ & $Acc_n\uparrow$ & $\Delta\downarrow$\\
\hline
Scenes & LwF \cite{lwf2017} & \textbf{71.0} & 69.9 & 52.3 & 1.7\\ 
$\rightarrow$ CUB & Ours & 70.7 & \textbf{69.9} &  \textbf{53.0} & \textbf{1.1}\\ \hline
CUB (150)  & LwF \cite{lwf2017} & 58.2 & 57.1 & 66.2 & 1.8 \\ 
$\rightarrow$ CUB (50) & Ours & \textbf{60.0} & \textbf{59.0} & \textbf{69.4} & \textbf{1.7}\\ 
\hline
\end{tabular}}
\end{minipage}

\caption{Experiment with (a) varying semantic representation and (b) 2D images.}
\label{table:sem}
\end{table}
\begin{figure*}[!t]
\centering
\vspace{-1em}\includegraphics[width=12cm]{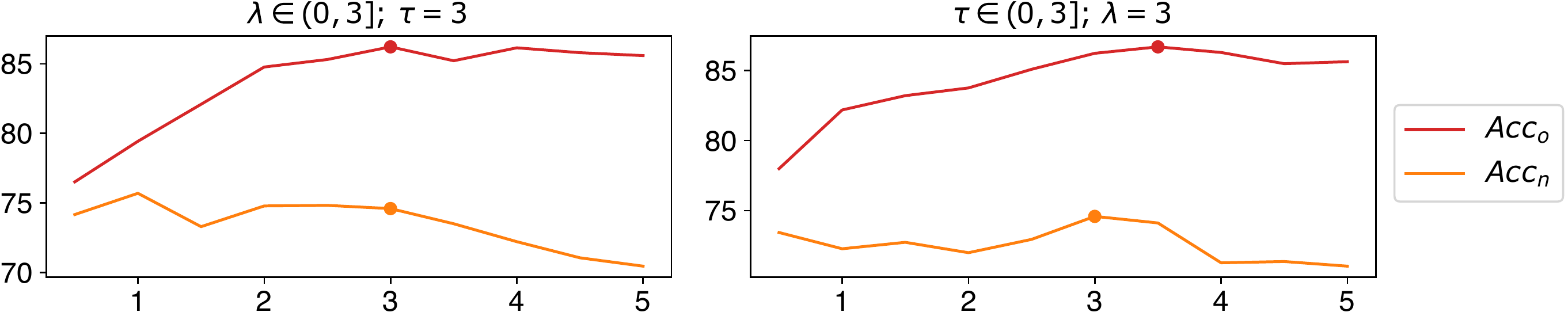}
\vspace{-1em}
\caption{Hyperparameter sensitivity on ModelNet40 $\rightarrow$ ModelNet10 settings.  Varying \textbf{(left)} $\lambda$ of Eq. \ref{eq:overall_loss} and \textbf{(right)} $\tau$ of $L_{KD}$ loss in Eq. \ref{eq:kd_loss}.}\vspace{-1em}
\label{fig:hyperparameter_sensitivity}
\end{figure*}

\noindent\textbf{Ablation studies:} In Table \ref{table:varying_backbone}, we perform ablation study while varying different 3D point cloud backbone. Among all backbones, PointNet performs consistently well in both 3D experiment settings. PointConv and DGCNN have some success in forgetting issue with synthetic data of ModelNet10 but fails to generalize it for real scanned ScanObjectNN classes. The global features extracted by PointNet may be more helpful than local features from PointConv and DGCNN backbones. Table \ref{table:sem}(a) also compares two different word vector models (word2vec and GloVe) as semantics. In most cases, word2vec achieves better accuracy and less forgetting in comparison to GloVe.

\noindent\textbf{Hyperparameter sensitivity:} We experiment with varying $\lambda$ and $\tau$ in Fig. \ref{fig:hyperparameter_sensitivity}. By fixing one hyperparameter and adjusting another, we observe hyperparameter sensitivity within the range $\lambda,\tau\in(0,3]$. We notice that increasing $\lambda$ and $\tau$ from 0 to 3 improves the old ($Acc_o$) and new ($Acc_n$) class performance. From $\lambda=\tau=3$ to higher, $Acc_o$ results do not deflect much, but $Acc_n$ values decrease gradually. We achieve best results using $\lambda=\tau=3$.

\begin{figure*}[!t]
\centering
\includegraphics[width=1\textwidth,trim={0cm 0cm 0cm 0cm},clip]{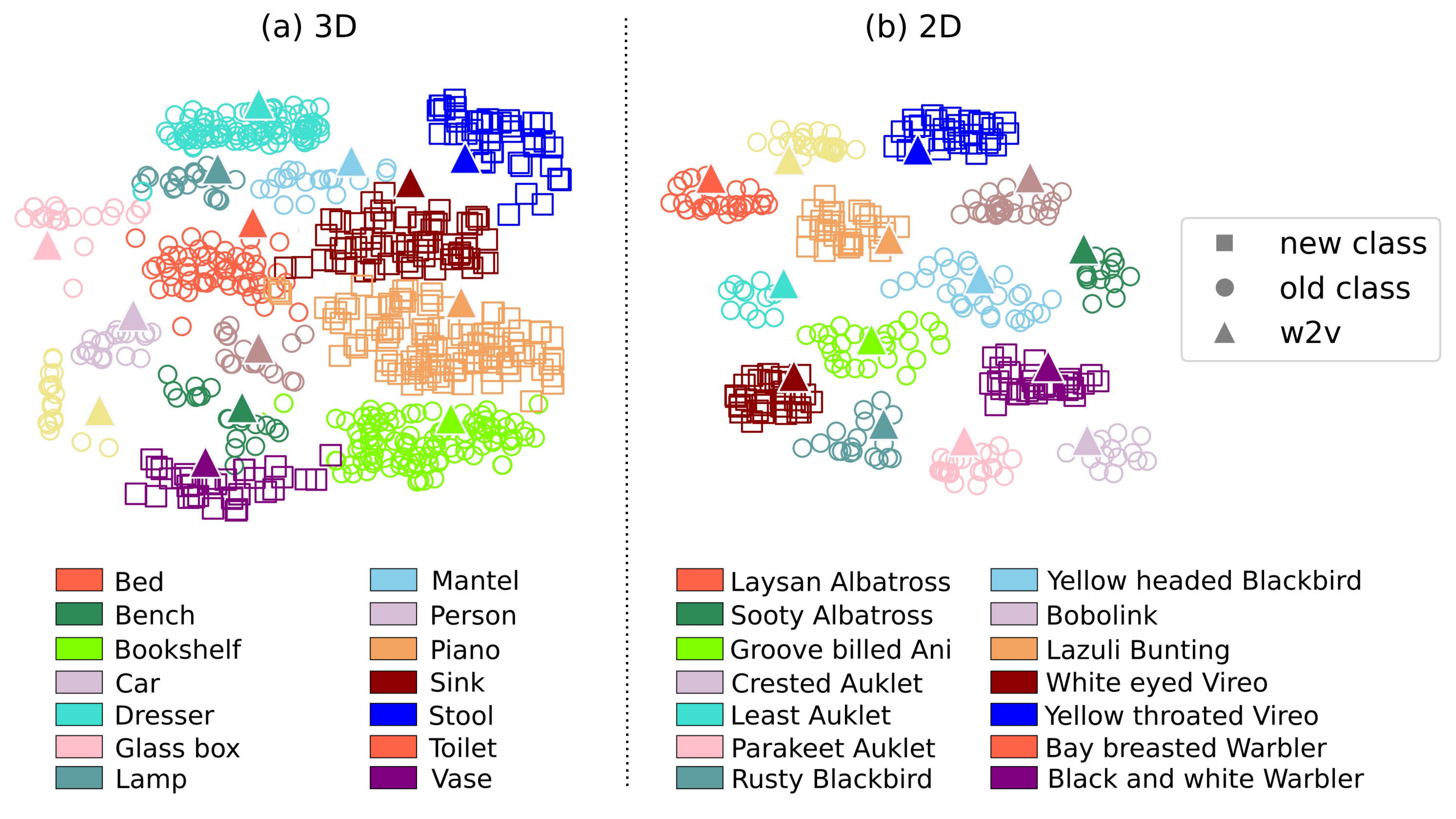}
\caption{tSNE visualization of features and semantics for (a) 2D image and (b) 3D point cloud objects. Ten old and four new classes are shown for better visualization. 2D image features are clustered better than 3D point cloud features.}
\label{fig:3Dvs2D}
\end{figure*}

\subsection{2D experiments}

In addition to 3D point cloud experiments, we conduct 2D image experiments. We report our results in Table \ref{table:sem}(b) using MIT Scenes \cite{scene2009}, CUB \cite{cub2011}. For two different experiment setups, Scenes $\rightarrow$ CUB and CUB (150) $\rightarrow$ CUB (50), our method achieves better performance than LwF \cite{lwf2017} in terms of less forgetting ($\Delta$). Moreover, we observe that the result of the 2D experiments is better than the 3D experiments (Tables \ref{table:overall_experiment_pointnet} and \ref{table:varying_backbone}). The amount of forgetting ($\Delta$) is higher in 3D cases than in 2D cases (5-6\% vs. 1-2\%). The main reason is the 2D backbone (VGG16 \cite{simonyan2014very}) has been pre-trained on a large dataset Imagenet \cite{imagenet2015}, which has million training instances and thousands of classes. In contrast, the 3D backbone (PointNet, DGCNN, PointConv) used in the 3D experiments is not pre-trained on a huge dataset. Therefore, the feature vector obtained from the 2D backbone is richer and more clustered than the feature vector obtained from the 3D backbone. We notice that the feature-semantic alignment in the 2D experiment is more aligned than the 3D experiment, as shown in Fig. \ref{fig:3Dvs2D}.

\section{Conclusion}
In this paper, we investigate LwF on 3D point cloud objects. Because of the lack of large-scale 3D datasets and powerful pre-trained models, popular knowledge distillation on prediction scores poorly performs on 3D data. To improve the performance further, we use semantic word vectors in the network pipeline. It helps to improve the traditional knowledge distillation performance. We also report performance on different 3D recognition backbones and word embeddings. We notice that the extent of forgetting in 3D is still inferior to the 2D image case. Future research in this area may investigate this issue further.

\noindent\textbf{Acknowledgment:} This work was supported by NSU CTRG 2020--2021 grant \#CTRG-20/SEPS/04.

%
%
%
\bibliographystyle{splncs04}
\bibliography{egbib}

\begin{thebibliography}{10}
\providecommand{\url}[1]{\texttt{#1}}
\providecommand{\urlprefix}{URL }
\providecommand{\doi}[1]{https://doi.org/#1}

\bibitem{aljundi2018memory}
Aljundi, R., Babiloni, F., Elhoseiny, M., Rohrbach, M., Tuytelaars, T.: Memory
  aware synapses: Learning what (not) to forget. In: Proceedings of the
  European Conference on Computer Vision (ECCV) (2018)

\bibitem{aljundi2017expert}
Aljundi, R., Chakravarty, P., Tuytelaars, T.: {Expert gate: Lifelong learning
  with a network of experts}. In: Proceedings of the IEEE/CVF Conference on
  Computer Vision and Pattern Recognition (CVPR) (2017)

\bibitem{castro2018}
Castro, F.M., Mar{\'\i}n-Jim{\'e}nez, M.J., Guil, N., Schmid, C., Alahari, K.:
  End-to-end incremental learning. In: Proceedings of the European Conference
  on Computer Vision (ECCV) (2018)

\bibitem{Ali_2021_CVPR}
Cheraghian, A., Rahman, S., Fang, P., Roy, S.K., Petersson, L., Harandi, M.:
  Semantic-aware knowledge distillation for few-shot class-incremental
  learning. In: Proceedings of the IEEE/CVF Conference on Computer Vision and
  Pattern Recognition (CVPR) (2021)

\bibitem{cheraghian2021zero}
Cheraghian, A., Rahman, S., Chowdhury, T.F., Campbell, D., Petersson, L.:
  Zero-shot learning on 3d point cloud objects and beyond. arXiv preprint
  arXiv:2104.04980  (2021)

\bibitem{empirical2014}
Goodfellow, I.J., Mirza, M., Xiao, D., Courville, A., Bengio, Y.: {An empirical
  investigation of catastrophic forgetting in gradient-based neural networks}.
  In: 2nd International Conference on Learning Representations, ICLR 2014 -
  Conference Track Proceedings (2014)

\bibitem{9127813}
Guo, Y., Wang, H., Hu, Q., Liu, H., Liu, L., Bennamoun, M.: Deep learning for
  3d point clouds: A survey. IEEE Transactions on Pattern Analysis and Machine
  Intelligence  (2020)

\bibitem{hinton2015}
Hinton, G., Vinyals, O., Dean, J.: Distilling the knowledge in a neural
  network. arXiv preprint arXiv:1503.02531  (2015)

\bibitem{hou2019learning}
Hou, S., Pan, X., Loy, C.C., Wang, Z., Lin, D.: {Learning a unified classifier
  incrementally via rebalancing}. In: Proceedings of the IEEE/CVF Conference on
  Computer Vision and Pattern Recognition (CVPR) (2019)

\bibitem{komarichev2019cnn}
Komarichev, A., Zhong, Z., Hua, J.: A-cnn: Annularly convolutional neural
  networks on point clouds. In: Proceedings of the IEEE/CVF Conference on
  Computer Vision and Pattern Recognition (CVPR) (2019)

\bibitem{li2018pointcnn}
Li, Y., Bu, R., Sun, M., Wu, W., Di, X., Chen, B.: Pointcnn: Convolution on
  x-transformed points. Advances in neural information processing systems
  (2018)

\bibitem{lwf2017}
Li, Z., Hoiem, D.: Learning without forgetting. IEEE Transactions on Pattern
  Analysis and Machine Intelligence  \textbf{40}(12),  2935--2947 (2018)

\bibitem{mallya2018packnet}
Mallya, A., Lazebnik, S.: Packnet: Adding multiple tasks to a single network by
  iterative pruning. In: Proceedings of the IEEE/CVF Conference on Computer
  Vision and Pattern Recognition (CVPR) (June 2018)

\bibitem{maturana2015voxnet}
Maturana, D., Scherer, S.: Voxnet: A 3d convolutional neural network for
  real-time object recognition. In: IROS (2015)

\bibitem{catastrophic1989}
McCloskey, M., Cohen, N.J.: {Catastrophic Interference in Connectionist
  Networks: The Sequential Learning Problem}. Psychology of Learning and
  Motivation - Advances in Research and Theory  (1989)

\bibitem{Mikolov_NIPS_2013}
Mikolov, T., Sutskever, I., Chen, K., Corrado, G.S., Dean, J.: Distributed
  representations of words and phrases and their compositionality. In: Burges,
  C.J.C., Bottou, L., Welling, M., Ghahramani, Z., Weinberger, K.Q. (eds.)
  Advances in Neural Information Processing Systems 26, pp. 3111--3119. Curran
  Associates, Inc. (2013)

\bibitem{ostapenko2019learning}
Ostapenko, O., Puscas, M., Klein, T., Jahnichen, P., Nabi, M.: {Learning to
  remember: A synaptic plasticity driven framework for continual learning}. In:
  Proceedings of the IEEE/CVF Conference on Computer Vision and Pattern
  Recognition (CVPR) (2019)

\bibitem{pennington2014glove}
Pennington, J., Socher, R., Manning, C.D.: Glove: Global vectors for word
  representation. In: EMNLP (2014)

\bibitem{pointnet2017}
Qi, C.R., Su, H., Mo, K., Guibas, L.J.: Pointnet: Deep learning on point sets
  for 3d classification and segmentation. In: Proceedings of the IEEE/CVF
  Conference on Computer Vision and Pattern Recognition (CVPR) (2017)

\bibitem{qi2016volumetric}
Qi, C.R., Su, H., Nie{\ss}ner, M., Dai, A., Yan, M., Guibas, L.J.: Volumetric
  and multi-view cnns for object classification on 3d data. In: Proceedings of
  the IEEE/CVF Conference on Computer Vision and Pattern Recognition (CVPR)
  (2016)

\bibitem{qi2017pointnet++}
Qi, C.R., Yi, L., Su, H., Guibas, L.J.: Pointnet++ deep hierarchical feature
  learning on point sets in a metric space. In: Proceedings of the 31st
  International Conference on Neural Information Processing Systems (2017)

\bibitem{scene2009}
Quattoni, A., Torralba, A.: Recognizing indoor scenes. In: Proceedings of the
  IEEE/CVF Conference on Computer Vision and Pattern Recognition (CVPR) (2009)

\bibitem{Rahman_2019_ICCV}
Rahman, S., Khan, S., Barnes, N.: Transductive learning for zero-shot object
  detection. In: Proceedings of the IEEE/CVF International Conference on
  Computer Vision (ICCV) (October 2019)

\bibitem{10.1007/978-3-030-69535-4_6}
Rahman, S., Khan, S., Barnes, N., Khan, F.S.: Any-shot object detection. In:
  Proceedings of the Asian Conference on Computer Vision (ACCV) (2020)

\bibitem{rahman2020IJCV}
Rahman, S., Khan, S.H., Porikli, F.: Zero-shot object detection: Joint
  recognition and localization of novel concepts. International Journal of
  Computer Vision  \textbf{128}(12),  2979--2999 (2020)

\bibitem{icarl2017}
Rebuffi, S.A., Kolesnikov, A., Sperl, G., Lampert, C.H.: {iCaRL: Incremental
  classifier and representation learning}. In: Proceedings of the IEEE/CVF
  Conference on Computer Vision and Pattern Recognition (CVPR) (2017)

\bibitem{imagenet2015}
Russakovsky, O., Deng, J., Su, H., Krause, J., Satheesh, S., Ma, S., Huang, Z.,
  Karpathy, A., Khosla, A., Bernstein, M., Berg, A.C., Fei-Fei, L.: {ImageNet
  Large Scale Visual Recognition Challenge}. International Journal on Computer
  Vision (IJCV)  (2015)

\bibitem{dgr}
Shin, H., Lee, J.K., Kim, J., Kim, J.: {Continual learning with deep generative
  replay}. In: Advances in Neural Information Processing Systems (2017)

\bibitem{simonovsky2017dynamic}
Simonovsky, M., Komodakis, N.: Dynamic edge-conditioned filters in
  convolutional neural networks on graphs. In: Proceedings of the IEEE/CVF
  Conference on Computer Vision and Pattern Recognition (CVPR) (2017)

\bibitem{simonyan2014very}
Simonyan, K., Zisserman, A.: Very deep convolutional networks for large-scale
  image recognition. In: Bengio, Y., LeCun, Y. (eds.) 3rd International
  Conference on Learning Representations, {ICLR} 2015, San Diego, CA, USA, May
  7-9, 2015, Conference Track Proceedings (2015)

\bibitem{su2015multi}
Su, H., Maji, S., Kalogerakis, E., Learned-Miller, E.: Multi-view convolutional
  neural networks for 3d shape recognition. In: Proceedings of the IEEE/CVF
  Conference on Computer Vision and Pattern Recognition (CVPR). pp. 945--953
  (2015)

\bibitem{scanobjectnn2019}
Uy, M.A., Pham, Q.H., Hua, B.S., Nguyen, T., Yeung, S.K.: {Revisiting point
  cloud classification: A new benchmark dataset and classification model on
  real-world data}. In: 2019 IEEE/CVF International Conference on Computer
  Vision ( ICCV) (2019)

\bibitem{cub2011}
Wah, C., Branson, S., Perona, P., Belongie, S.: {Multiclass recognition and
  part localization with humans in the loop}. In: International Conference on
  Computer Vision ( ICCV) (2011)

\bibitem{wang2017cnn}
Wang, P.S., Liu, Y., Guo, Y.X., Sun, C.Y., Tong, X.: O-cnn: Octree-based
  convolutional neural networks for 3d shape analysis. ACM Transactions on
  Graphics (TOG)  (2017)

\bibitem{wang2019dynamic}
Wang, Y., Sun, Y., Liu, Z., Sarma, S.E., Bronstein, M.M., Solomon, J.M.:
  Dynamic graph cnn for learning on point clouds. Acm Transactions On Graphics
  (tog)  (2019)

\bibitem{wu2018memory}
Wu, C., Herranz, L., Liu, X., Wang, Y., Van De~Weijer, J., Raducanu, B.:
  {Memory replay Gans: Learning to generate images from new categories without
  forgetting}. In: Advances in Neural Information Processing Systems (2018)

\bibitem{pointconv2019}
Wu, W., Qi, Z., Fuxin, L.: {PointCONV: Deep convolutional networks on 3D point
  clouds}. In: Proceedings of the IEEE/CVF Conference on Computer Vision and
  Pattern Recognition (CVPR) (2019)

\bibitem{modelnet2015}
Wu, Z., Song, S., Khosla, A., Yu, F., Zhang, L., Tang, X., Xiao, J.: {3D
  ShapeNets: A deep representation for volumetric shapes}. In: Proceedings of
  the IEEE/CVF Conference on Computer Vision and Pattern Recognition (CVPR)
  (2015)

\bibitem{xian2018zero}
Xian, Y., Lampert, C.H., Schiele, B., Akata, Z.: Zero-shot learning—a
  comprehensive evaluation of the good, the bad and the ugly. IEEE transactions
  on pattern analysis and machine intelligence  (2018)

\bibitem{yang2019learning}
Yang, Z., Wang, L.: Learning relationships for multi-view 3d object
  recognition. In: Proceedings of the IEEE/CVF Conference on Computer Vision
  and Pattern Recognition (CVPR) (2019)

\bibitem{Zhang2020}
Zhang, J., Zhang, J., Ghosh, S., Li, D., Tasci, S., Heck, L., Zhang, H., Kuo,
  C.C.J.: Class-incremental learning via deep model consolidation. In: Workshop
  on Applications of Computer Vision (WACV) (2020)

\bibitem{Semantic_Relation_2021_CVPR}
Zhu, C., Chen, F., Ahmed, U., Savvides, M.: Semantic relation reasoning for
  shot-stable few-shot object detection. arXiv preprint arXiv:2103.01903
  (2021)

\end{thebibliography}

\end{document}